\documentclass[runningheads]{llncs}
\usepackage[T1]{fontenc}
\usepackage{graphicx}
\usepackage[misc]{ifsym} 
\usepackage{amssymb}
\usepackage{amsmath}
\usepackage{multirow}
\usepackage{booktabs}
\usepackage{color}
\usepackage[colorlinks,linkcolor=blue]{hyperref}
\newcommand{\repeatthanks}{\textsuperscript{\thefootnote}}

\begin{document}
\title{Multi-IMU with Online Self-Consistency for Freehand 3D Ultrasound Reconstruction}

\author{
Mingyuan Luo\inst{1,2,3}\thanks{Mingyuan Luo and Xin Yang contribute equally to this work.} \and
Xin Yang\inst{1,2,3}\repeatthanks \and
Zhongnuo Yan\inst{1,2,3} \and
Junyu Li\inst{1,2,3} \and
Yuanji Zhang\inst{1,2,3} \and
Jiongquan Chen\inst{1,2,3} \and
Xindi Hu\inst{4} \and
Jikuan Qian\inst{4} \and
Jun Cheng\inst{1,2,3} \and
Dong Ni\inst{1,2,3}\textsuperscript{(\Letter)}
}


\authorrunning{M. Luo et al.}
\titlerunning{Multi-IMU with Online Self-Consistency for Freehand 3D US Reconstruction}

\institute{
National-Regional Key Technology Engineering Laboratory for Medical Ultrasound, School of Biomedical Engineering, Health Science Center, Shenzhen University, China \\
\email{nidong@szu.edu.cn} \and
Medical Ultrasound Image Computing (MUSIC) Lab, Shenzhen University, China \and
Marshall Laboratory of Biomedical Engineering, Shenzhen University, China \and
Shenzhen RayShape Medical Technology Inc., Shenzhen, China
}

\maketitle              

\begin{abstract}
Ultrasound (US) imaging is a popular tool in clinical diagnosis, offering safety, repeatability, and real-time capabilities. Freehand 3D US is a technique that provides a deeper understanding of scanned regions without increasing complexity. However, estimating elevation displacement and accumulation error remains challenging, making it difficult to infer the relative position using images alone. The addition of external lightweight sensors has been proposed to enhance reconstruction performance without adding complexity, which has been shown to be beneficial. We propose a novel online self-consistency network (OSCNet) using multiple inertial measurement units (IMUs) to improve reconstruction performance. OSCNet utilizes a modal-level self-supervised strategy to fuse multiple IMU information and reduce differences between reconstruction results obtained from each IMU data. Additionally, a sequence-level self-consistency strategy is proposed to improve the hierarchical consistency of prediction results among the scanning sequence and its sub-sequences. Experiments on large-scale arm and carotid datasets with multiple scanning tactics demonstrate that our OSCNet outperforms previous methods, achieving state-of-the-art reconstruction performance.
\end{abstract}

\begin{figure}
\centering
\includegraphics[width=\textwidth]{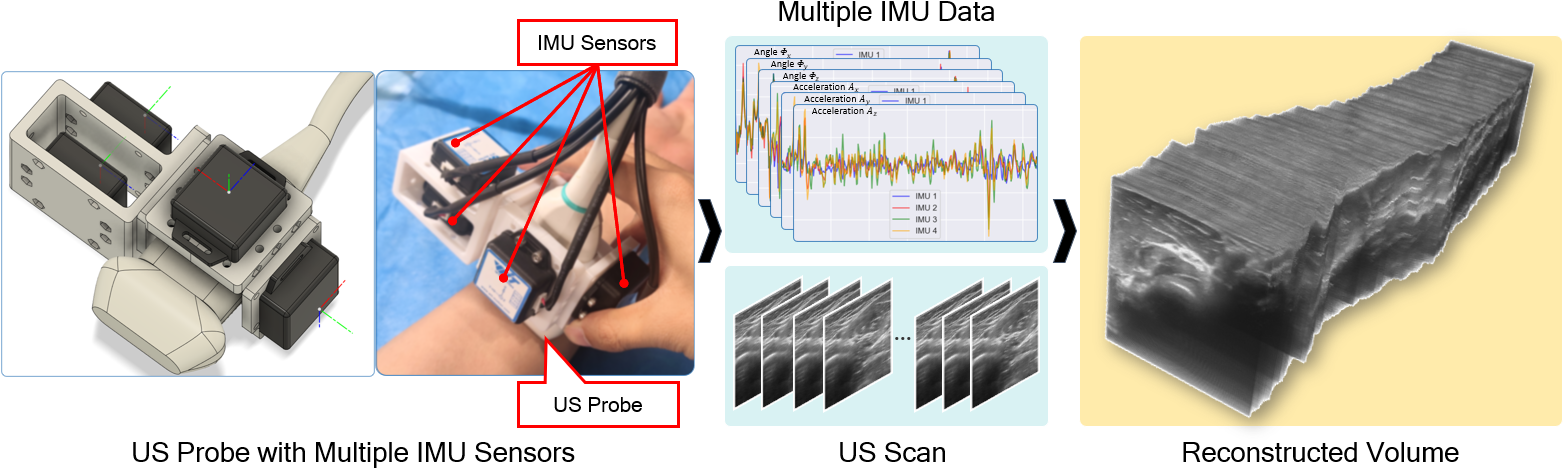}
\caption{Pipeline of freehand 3D US reconstruction with multiple lightweight inertial measurement unit (IMU) sensors.} \label{fig:pipeline}
\end{figure}

\section{Introduction}
Ultrasound (US) imaging has been widely used in clinical diagnosis due to its advantages of safety, repeatability, and real-time imaging. Compared with 2D US, 3D US can provide more comprehensive spatial information. Freehand 3D US can enhance the understanding of physicians about the scanned region of interest without increasing the complexity of scanning~\cite{Siang19,PREVOST2018187,MingyuanLuo2022DeepMN}. However, the difficulty in estimating elevation displacement and accumulation error makes it very challenging to infer the relative position only from images. In this regard, it is expected to improve the reconstruction performance with the help of external lightweight sensors, which will not significantly increase the scanning complexity.

Sensorless freehand 3D US reconstructs the volume by calculating the relative transformation of a series of US images. Previous studies were mainly based on speckle decorrelation~\cite{JianFengChen1997DeterminationOS,TheresaATuthill1998AutomatedTU}, which estimates out-of-plane motion through the correlation of speckle patterns in two successive frames. With the development of deep learning technology, recent studies were mainly based on convolutional neural network (CNN). Prevost et al.~\cite{RaphaelPrevost2017DeepLF} proposed an end-to-end method based on CNN to estimate the relative motion of US images. Guo et al.~\cite{HengtaoGuo2020SensorlessF3} proposed a deep contextual learning network (DCL-Net) based on 3D CNN to estimate the trajectory of US probe, and in a more recent study~\cite{HengtaoGuo2022UltrasoundVR}, they proposed a deep contextual-contrastive network (DC$^2$-Net), which introduced a contrastive learning strategy to improve the reconstruction performance by leveraging the label efficiently. Luo et al.~\cite{MingyuanLuo2021SelfCA,MingyuanLuo2023RecON} proposed an online learning framework (OLF) that improves reconstruction performance by online learning and shape priors.

Due to the low cost, small size, and low power consumption of micro-electro-mechanical-systems (MEMS), the sensor called inertial measurement unit (IMU) has been widely used in navigation systems. Prevost et al.~\cite{PREVOST2018187} incorporated IMU orientation into neural network to improve reconstruction performance. Luo et al.~\cite{MingyuanLuo2022DeepMN} proposed a deep motion network (MoNet) to mine the valuable information of low signal-to-noise acceleration, and an online self-supervised strategy was designed to further improve reconstruction performance. However, the main disadvantage of IMU is that its measurement noise can not be completely eliminated by calibration. Existing studies have shown that combining multiple IMUs may help reduce noise and improve accuracy~\cite{guerrier2009improving,7483205}.

In this study, we propose a multi-IMU-based online self-consistency network (OSCNet) for freehand 3D US reconstruction. Our contribution is two-fold. First, we equip multiple IMUs (see Fig.~\ref{fig:pipeline}) to reduce the influence of noise in individual IMU data. We propose a modal-level self-supervised strategy (MSS) to fuse the information from multiple IMUs. MSS improves reconstruction performance by reducing the differences between reconstruction results obtained from each IMU data. Second, to reduce the estimation instability caused by scanning differences such as frame rates, we propose a sequence-level self-consistency strategy (SCS), which improves the hierarchical consistency of prediction results among the scanning sequence and its sub-sequences based on a consistent context. Experimental results show that the proposed OSCNet can effectively fuse the information of multiple IMUs and achieve state-of-the-art reconstruction performance.

\section{Methodology}

\begin{figure}[t]
\centering
\includegraphics[width=1.0\textwidth]{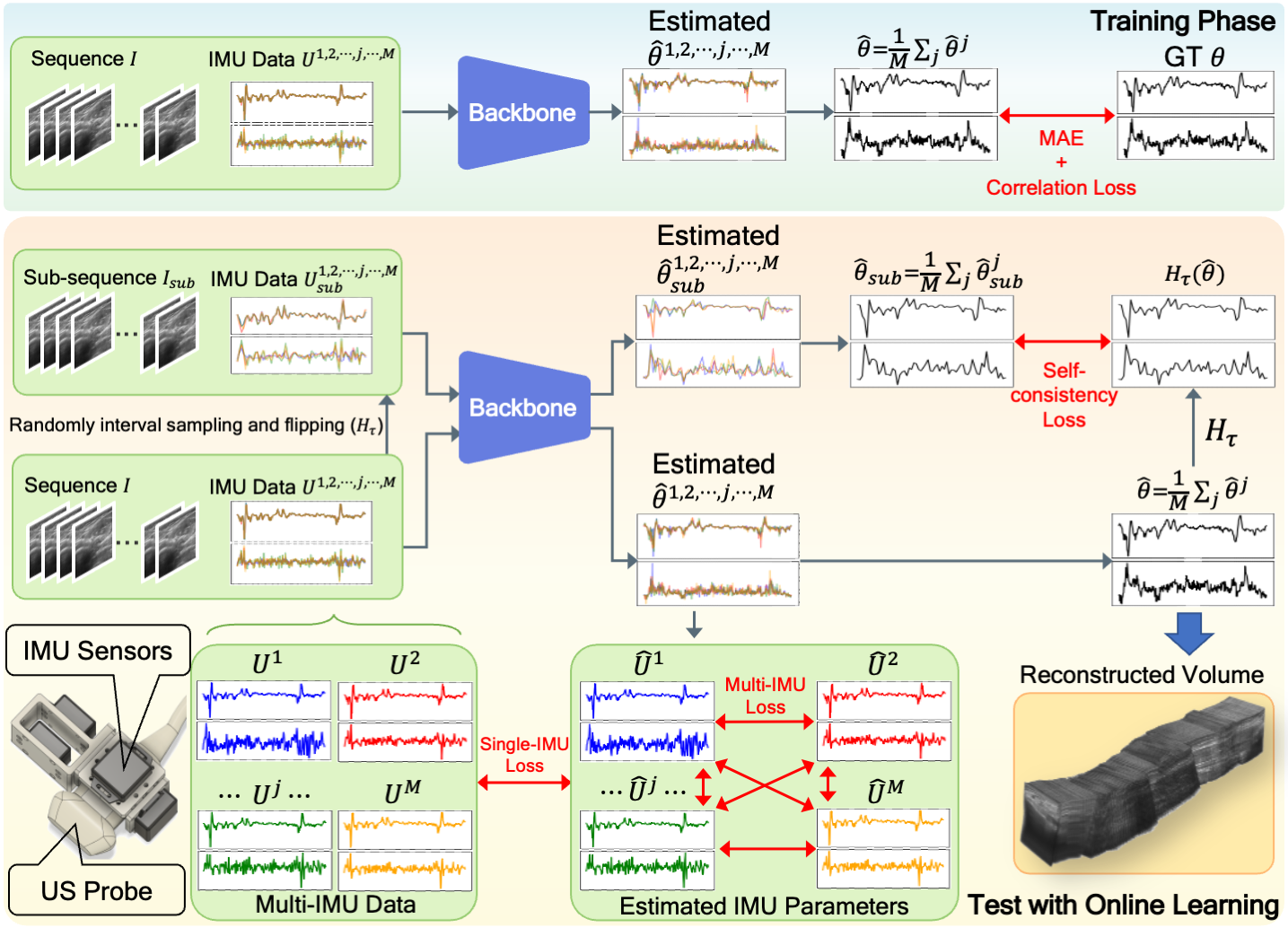}
\caption{Overview of our proposed multi-IMU online self-consistency network (OSCNet). IMU data diagrams ($U/\hat{U}$) show angle curves ($\Phi/\hat{\Phi}$, top) and acceleration curves ($A/\hat{A}$, bottom). Relative transformation parameter diagrams ($\theta/\hat{\theta}$) show angle curves ($\phi/\hat{\phi}$, top) and translation curves ($t/\hat{t}$, bottom).}
\label{fig:framework}
\end{figure}

Fig.~\ref{fig:framework} illustrates the proposed OSCNet, which consists of two essential components: backbone and online learning. We construct a backbone using the temporal and multi-branch structure from~\cite{MingyuanLuo2022DeepMN}. The main branch in the backbone consists of ResNet~\cite{he2016deep} for feature extraction and LSTM~\cite{hochreiter1997long} for processing temporal information. It aids future estimation by leveraging temporal contextual information. Additionally, there is an independent motion branch in the backbone that fuses IMU information from a motion perspective with US images. For more details, please refer to~\cite{MingyuanLuo2022DeepMN}.

In the training phase, we input an $N$-length scanning sequence $I=\{I_i|i=1,2,\cdots,N\}$ and corresponding multiple IMU data $U=\{U_i|i=1,2,\cdots,N-1\}$ into the backbone to estimate the 3D relative transformation parameters $\theta=\{\theta_i|i=1,2,\cdots,N-1\}$, where $\theta_i$ includes 3-axis translations $t_i=(t_x,t_y,t_z)_i$ and rotation angles $\phi_i=(\phi_x,\phi_y,\phi_z)_i$ between image $I_i$ and $I_{i+1}$. The multiple IMU data consists of $M$ independent IMU data $U_i=\{U^j_i|j=1,2,\cdots,M\}$, where $U^j_i$ consists of 3-axis angles $\Phi^j_i=(\Phi_x,\Phi_y,\Phi_z)^j_i$ and accelerations $A^j_i=(A_x,A_y,A_z)^j_i$. The pre-processing process for $\Phi_i$ and $A_i$ is consistent with~\cite{MingyuanLuo2022DeepMN}.
Compared to traditional offline inference strategies, online learning can leverage valuable information from unlabeled data to improve the model's performance~\cite{MingyuanLuo2021SelfCA,MingyuanLuo2022DeepMN}. In the testing phase, we propose two online self-supervised strategies based on both the multiple IMU data (modal-level) and the scanning sequence itself (sequence-level) to improve the performance of the backbone's estimations.

\subsection{Modal-level Self-supervised Strategy}
Multiple IMUs mounted in different directions provide diverse measurement constraints for the model's estimation, as shown in Fig.~\ref{fig:pipeline}. This makes it possible to reduce the influence of noise in individual IMU data while adaptively optimizing for estimation. We construct an online modal-level self-supervised strategy (MSS) that leverages the consistency between the backbone's estimation and multiple IMU data to improve the reconstruction performance.

As shown in the top of Fig.~\ref{fig:framework}, during the training phase, we repeatedly input the US images and $M$ different IMU data into the backbone to obtain $M$ estimated parameters. We use the average of the $M$ estimated parameters $\hat{\theta}=\frac{1}{M}\sum^M_{j=1}\hat{\theta}^j$ as the final output of the backbone. We then calculate training loss between $\hat{\theta}$ and ground truth $\theta$ using mean absolute error (MAE) and Pearson correlation loss~\cite{HengtaoGuo2020SensorlessF3}:
\begin{equation}
L=\|\hat\theta-\theta\|_1+(1-\frac{\textbf{Cov}(\hat\theta,\theta)}{\sigma(\hat\theta)\sigma(\theta)}),
\end{equation}
where $\textbf{Cov}$, $\sigma$ and $\left\|\cdot\right\|_1$ denote the covariance, the standard deviation, and L1 normalization, respectively.

As shown in the bottom of Fig.~\ref{fig:framework}, during the testing phase, we use each IMU data $U^j$ ($j=1,2,\cdots,M$) as a weak label to constrain the corresponding estimated parameters $\hat{\theta}^j$. We calculate the estimated acceleration $\hat{A}^j$ at the center point of each image using the estimated $\hat{\theta}^j$. To reduce the influence of acceleration noise, we scale the $\hat{A}^j$ to match the mean-zeroed IMU acceleration.
\begin{equation}
    \hat{A}^j_i=((\hat{t}^j_{i-1})^{-1}+\hat{t}^j_{i})-\frac{1}{N-2}\sum_i((\hat{t}^j_{i-1})^{-1}+\hat{t}^j_{i}),\quad i=2,3,\cdots,N-1,
\end{equation}
where $(\hat{t}^j_{i-1})^{-1}$ represents the translations in the inversion of $\hat{\theta}^j_{i-1}$. Similar to~\cite{MingyuanLuo2022DeepMN}, we use Pearson correlation loss to measure the difference between the estimated and IMU acceleration, while the angle is measured using MAE. Therefore, the single-IMU consistency constraint between the estimated parameters and corresponding IMU data can be expressed as:
\begin{equation}
L_{single-IMU}=\sum^M_{j=1}(1-\frac{\textbf{Cov}(\hat{A}^j,A^j)}{\sigma(\hat{A}^j)\sigma(A^j)})+\|\hat\phi^j-\Phi^j\|_1.
\end{equation}

In addition, the consistency among multiple IMU data itself also provides the possibility to improve the reconstruction performance. It constrains the backbone to obtain similar estimated parameters for different IMU data inputs from the same scan. Specifically, we construct multi-IMU consistency constraints as:
\begin{equation}
L_{multi-IMU}=\sum^M_{j,k=1,2,\cdots,M, j<k}(1-\frac{\textbf{Cov}(\hat{A}^j,\hat{A}^k)}{\sigma(\hat{A}^j)\sigma(\hat{A}^k)})+\|\hat\phi^j-\hat\phi^k\|_1.
\end{equation}

\subsection{Sequence-level Self-consistency Strategy}
Consistent context should lead to consistent parameter estimation, which constrains the model at the sequence level, reducing the estimation instability caused by scanning differences such as frame rates. Inspired by contrastive learning~\cite{9226466}, we construct an online sequence-level self-consistency strategy (SCS). SCS randomly generates sub-sequences with consistent context for each scan. The hierarchical consistency constraint among the generated sub-sequences and the original sequence improves the reconstruction performance of the backbone. Specifically, as shown in Fig.~\ref{fig:framework}, we randomly interval sample and flip each scanning sequence $I$ and its IMU data $U$ to generate a sub-sequence $I_{sub}$ ($U_{sub}$) with consistent context. In the testing phase, we obtain the estimated parameters $\hat{\theta}_{sub}$ of $I_{sub}$ ($U_{sub}$) using the trained backbone. Then compare $\hat{\theta}_{sub}$ with the original estimated parameters $\hat{\theta}$ after the same interval sampling and flipping to construct the self-consistency constraint:
\begin{equation}
\begin{split}
L_{self-consistency}=&\|\hat\theta_{sub}-H_\tau(\hat\theta)\|_1\\
=&\|\frac{1}{M}\sum^M_{j=1}B(H_\tau(I),H_\tau(U^j))-H_\tau(\frac{1}{M}\sum^M_{j=1}B(I, U^j))\|_1,
\end{split}
\end{equation}
where $H_\tau$ converts the parameters, sequences, or IMU data under interval sampling and flipping operation $\tau$. $B$ denotes the backbone.

\section{Experiments}

\begin{figure}[!h]
\centering
\includegraphics[width=1.0\textwidth]{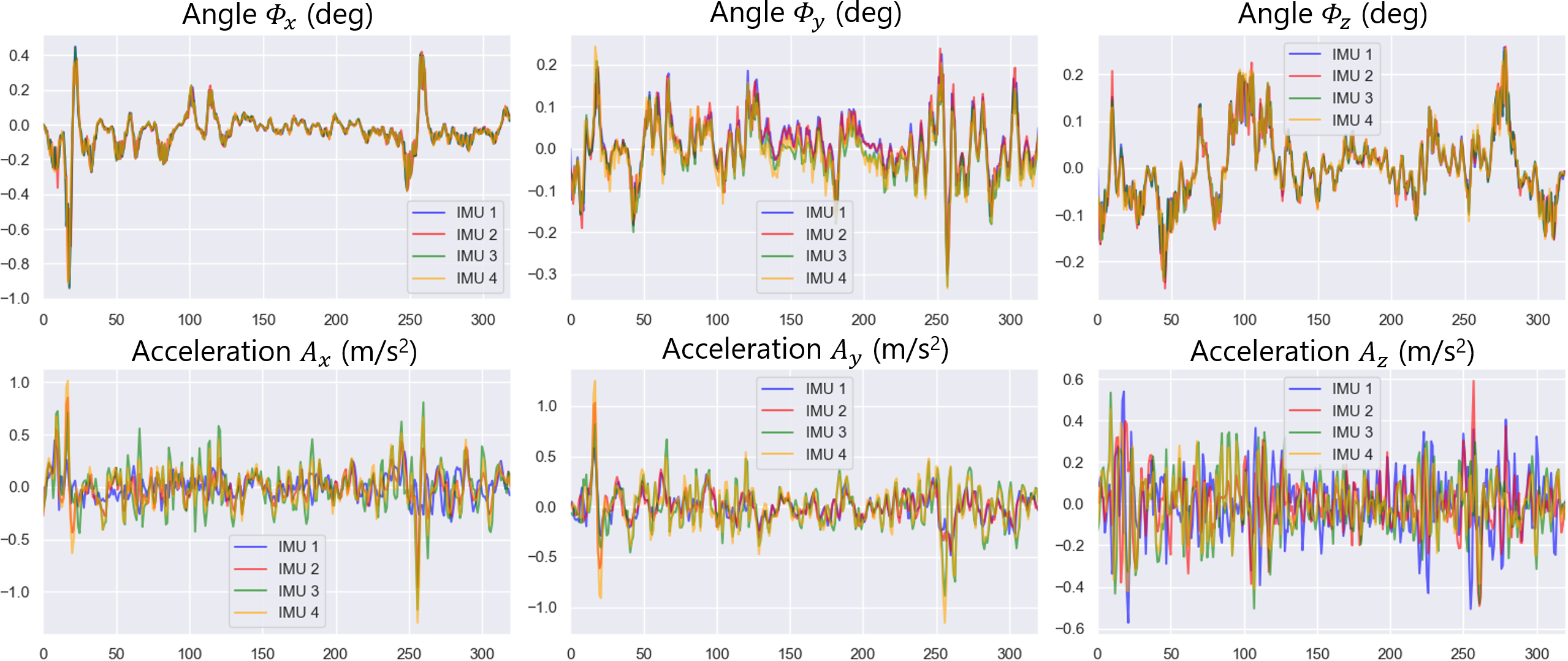}
\caption{Comparison of multiple IMU data. The abscissa of each subfigure indicates the image index.}
\label{fig:imu_cali_result}
\end{figure}

\paragraph{\textbf{Materials and Implementation.}} The equipment we used to collect data includes a portable US machine, four IMU sensors (WT901C-232, WitMotion) and an electromagnetic (EM) positioning system. The US images were acquired with a linear probe at 10 MHz, and the depth was set at 4 cm. As shown in Fig.~\ref{fig:pipeline}, we bound four IMU sensors to the probe in different orientations (three for 3-axis directions and one for redundancy) using a 3D-printed bracket, which can compensate for errors and reduce measurement singularities~\cite{mi13081214}. The resolutions of the IMU acceleration and angle are $5\times10^{-4}$ g/LSB and 0.5$^\circ$, respectively. We used the EM positioning system to trace the scan route accurately. The direction and angle resolutions of the EM positioning system are 1.4 mm and 0.5$^\circ$, respectively. We calibrated the multiple IMU sensors and the EM positioning system using the Levenberg–Marquardt algorithm~\cite{levenberg1944method} to ensure accurate measurements and minimise system errors. As shown in Fig.~\ref{fig:imu_cali_result}, the calibrated IMU data exhibits a generally consistent overall trend, although differences still exist.

We constructed two datasets, including arm and carotid, from 36 volunteers. The arm dataset contains 288 scans, with scanning tactics including linear, curved, loop, and sector scan. The carotid dataset contains 216 scans, with scanning tactics including linear, loop, and sector scan. The average lengths of the arm and carotid scans are 323.96 mm and 203.25 mm, respectively. The size of scanned images is $248\times260$ pixels, and the image spacing is $0.15\times0.15$ mm$^2$. The collection and use of the data are approved by the local IRB.

The arm and carotid datasets were randomly divided into 200/40/48 and 150/30/36 scans based on volunteer level to construct training/validation/test set. To prevent overfitting and enhance the model's robustness, we performed random augmentations on each scan, including sub-sequence intercepting, interval sampling, and sequence inversion. We randomly augmented each training scan to 20 sequences and each test scan to 10 sequences to simulate complex real-world situations. We used the Adam optimizer to optimize the OSCNet. During the training phase, the epochs and batch size are set to 200 and 1, respectively. To avoid overfitting, we set the initial learning rate to $2\times 10^{-4}$ and used a learning rate decay strategy that halves the learning rate every 30 epochs. During the online learning phase, the iteration epoch and learning rate are set to 60 and $2\times 10^{-6}$, respectively. All code was implemented in PyTorch and executed on an RTX 3090 GPU.

\begin{table}[!h]
\caption{The mean (std) results of different models on the arm and carotid scans. DC$^2$: DC$^2$-Net, Bk: Backbone. The best results are shown in blue.}
\resizebox{\textwidth}{!}{
\begin{tabular}{l|c|c|c|c|c|c}
\toprule
\textbf{Models} & \textbf{FDR(\%)$\downarrow$} & \textbf{ADR(\%)$\downarrow$} & \textbf{MD(mm)$\downarrow$} & \textbf{SD(mm)$\downarrow$} & \textbf{HD(mm)$\downarrow$} & \textbf{EA(deg)$\downarrow$} \\ 
\hline
&\multicolumn{6}{c}{Arm scans} \\
\hline
CNN~\cite{PREVOST2018187} & 23.31(13.0) & 32.54(13.7) & 67.79(30.0) & 2313.08(1852.5) & 62.48(31.5) & 4.35(1.8) \\
DC$^2$~\cite{HengtaoGuo2022UltrasoundVR} & 14.02(7.3) & 26.15(10.2) & 45.50(21.3) & 1560.30(1181.4) & 42.25(40.9) & 4.71(2.3) \\
MoNet~\cite{MingyuanLuo2022DeepMN} & 11.58(6.2) & 20.35(6.8) & 32.21(11.0) & 1205.42(742.6) & 31.03(11.3) & 3.98(1.5) \\
Bk & 13.32(8.2) & 23.21(9.6) & 36.39(13.6) & 1339.17(822.4) & 34.91(13.7) & 4.32(1.7) \\
Bk+MSS & 10.78(5.6) & 19.53(6.3) & 30.52(10.5) & 1142.42(636.8) & 29.32(10.8) & 3.18(1.4) \\
Bk+SCS & 10.56(5.9) & 19.57(6.6) & 29.84(11.1) & 1126.28(614.9) & 28.64(11.6) & 3.65(1.9) \\
OSCNet & \textcolor{blue}{10.01(5.7)} & \textcolor{blue}{18.86(6.5)} & \textcolor{blue}{28.61(11.0)} & \textcolor{blue}{1064.06(582.5)} & \textcolor{blue}{27.38(11.4)} & \textcolor{blue}{2.76(1.3)} \\
\hline
&\multicolumn{6}{c}{Carotid scans} \\
\hline
CNN~\cite{PREVOST2018187} & 25.85(15.0) & 33.95(16.8) & 49.64(25.5) & 1944.72(1485.1) & 39.30(18.7) & 3.73(2.3) \\
DC$^2$~\cite{HengtaoGuo2022UltrasoundVR} & 13.54(7.1) & 21.68(9.2) & 26.47(9.6) & 1025.06(622.8) & 24.49(10.3) & 4.30(3.0) \\
MoNet~\cite{MingyuanLuo2022DeepMN} & 11.80(5.7) & 20.42(8.8) & 23.48(8.7) & 894.39(381.0) & 20.78(9.2) & 3.67(2.1) \\
Bk & 12.85(6.5) & 21.78(10.5) & 24.65(9.1) & 965.12(466.6) & 21.81(9.5) & 3.83(2.0) \\
Bk+MSS & 11.31(5.4) & 20.04(8.8) & 22.72(8.1) & 850.68(321.7) & 20.02(8.5) & 3.16(1.8) \\
Bk+SCS & 11.30(5.4) & 20.16(8.6) & 23.01(8.4) & 863.48(320.9) & 20.56(8.7) & 3.36(1.8) \\
OSCNet & \textcolor{blue}{10.90(5.3)} & \textcolor{blue}{19.61(8.5)} & \textcolor{blue}{21.81(7.2)} & \textcolor{blue}{804.27(282.8)} & \textcolor{blue}{19.30(7.6)} & \textcolor{blue}{2.60(1.6)} \\
\bottomrule
\end{tabular}
}
\label{tab:armcarotid}
\end{table}

\begin{figure}[!h]
\centering
\includegraphics[width=0.98\textwidth]{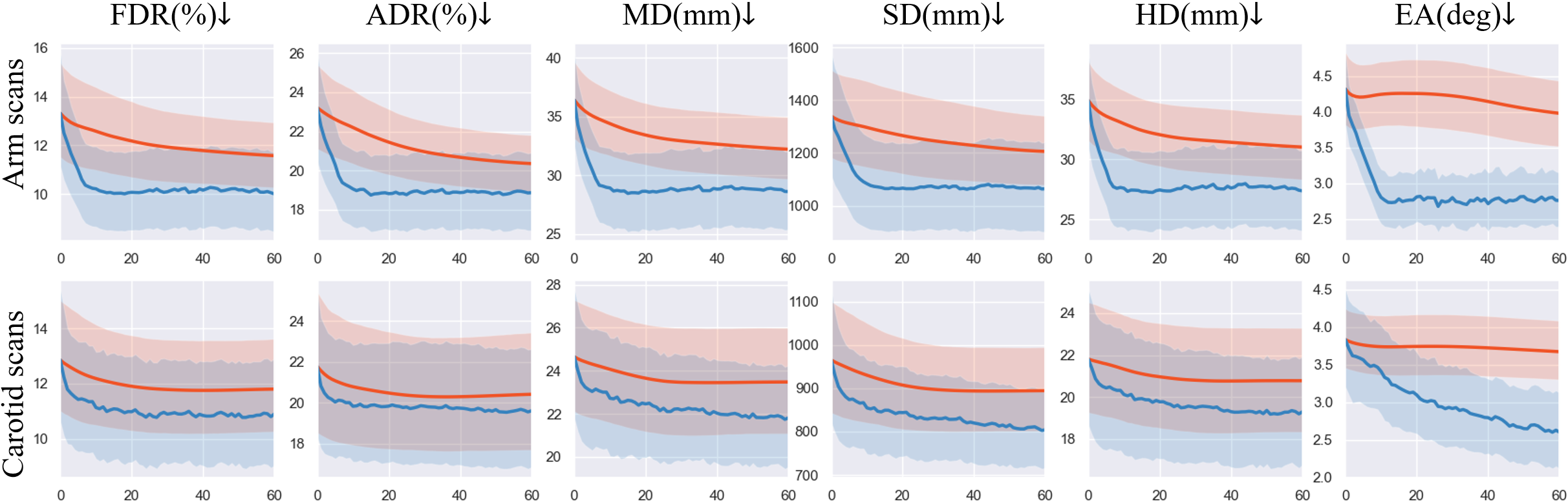}
\caption{Metric decline curves (with 95\% confidence interval). Red: MoNet, Blue: OSCNet. The abscissa and ordinate of each subfigure represent the number of iterations and the value of metrics, respectively.}
\label{fig:curve}
\end{figure}

\paragraph{\textbf{Quantitative and Qualitative Analysis.}} To demonstrate the effectiveness of our OSCNet, we compared it with three state-of-the-art methods, including CNN~\cite{PREVOST2018187}, DC$^2$-Net~\cite{HengtaoGuo2022UltrasoundVR} and MoNet~\cite{MingyuanLuo2022DeepMN}. All comparison methods were trained to convergence using the experimental settings given in the corresponding papers. We adopt six metrics from~\cite{MingyuanLuo2022DeepMN} to evaluate reconstruction performance: final drift rate (FDR), average drift rate (ADR), maximum drift (MD), sum of drift (SD), symmetric Hausdorff distance (HD), and mean error of angle (EA). In addition, ablation experiments are conducted to validate the effectiveness of MSS and SCS as proposed in our OSCNet.

Table~\ref{tab:armcarotid} shows that our OSCNet significantly outperforms CNN, DC$^2$-Net, MoNet, and our Backbone in all metrics for both arm and carotid scans ($t$-test, $p<0.05$), except for MoNet's ADR on the carotid scans ($t$-test, $p=0.10$). Notably, sensor-based methods (MoNet and OSCNet) have exhibited improvements in all metrics compared to sensorless methods (DC$^2$-Net and CNN). The multi-IMU-based OSCNet outperforms the single-IMU-based MoNet, verifying the effectiveness of multiple IMU integration. Moreover, the ablation experiments further demonstrate that both multiple IMU integration (MSS) and self-consistency constraint (SCS) greatly improve the reconstruction performance.

\begin{figure}[t]
\centering
\includegraphics[width=\textwidth]{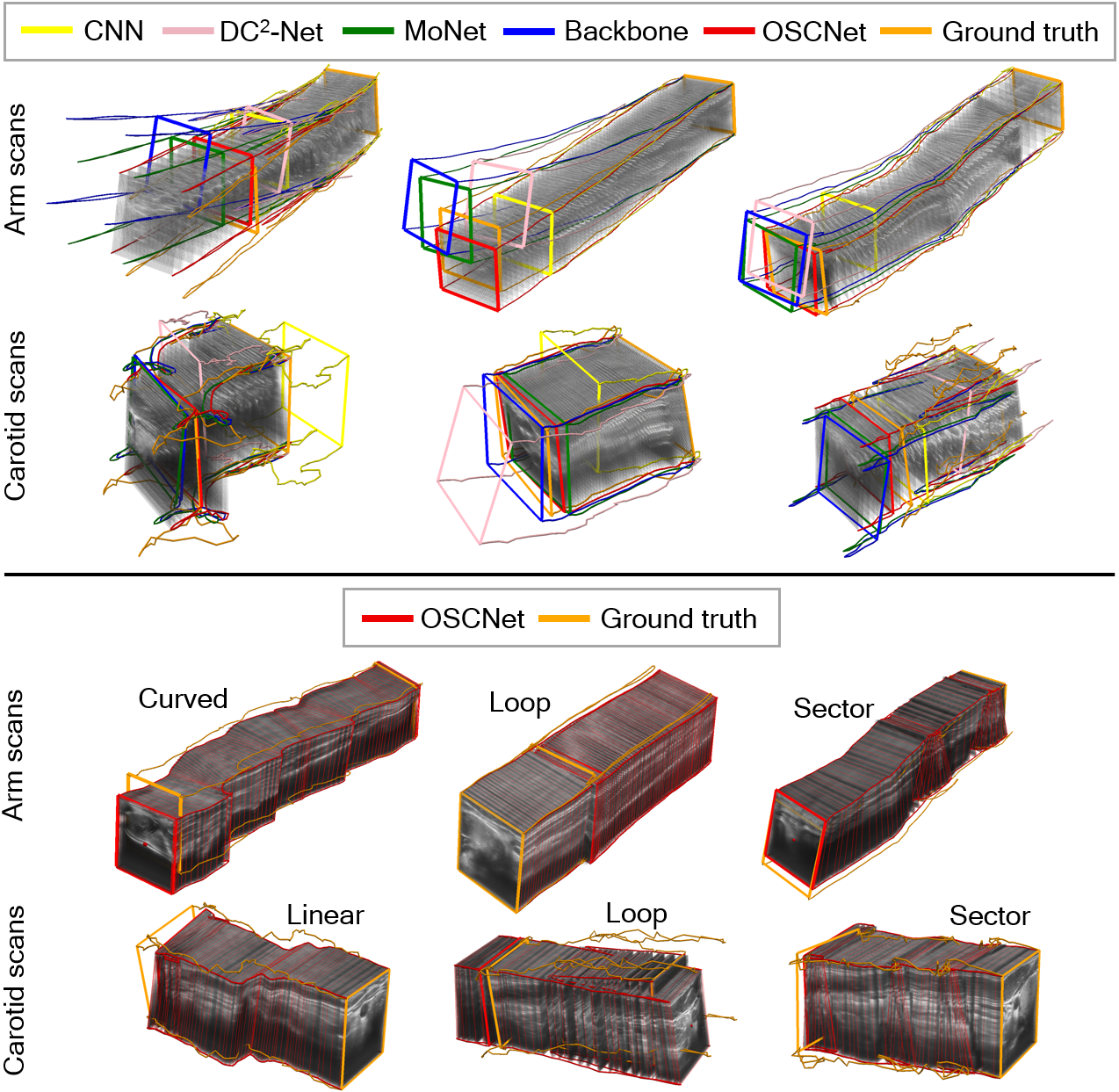}
\caption{Typical reconstruction cases. Top: comparison of different methods, Bottom: comparison of different scanning tactics. At the bottom, all of the estimated image positions of OSCNet are marked with red boxes to visualize the scanning tactics.}
\label{fig:reco}
\end{figure}

In addition, Fig.~\ref{fig:curve} displays the metric decline curves during the online learning phase of MoNet and OSCNet on the arm and carotid datasets. All metric curves exhibit a decreasing trend followed by stabilization. We note that our OSCNet has achieved further improvements compared to MoNet, with 13.56\% /7.32\%/30.65\% and 7.62\%/4.00\%/29.16\% improvement in FDR/ADR/EA on the arm and carotid datasets, respectively.
Fig.~\ref{fig:reco} presents several typical reconstruction results of all methods. It can be observed that our OSCNet outperforms other methods in reconstruction results and closely approximates the ground truth across all scanning tactics.

\section{Conclusion}

In this study, we propose a novel multi-IMU-based online self-consistency network (OSCNet) to conduct freehand 3D US reconstruction. We propose an online modal-level self-supervised strategy (MSS) that integrates multiple IMUs to reduce the influence of single IMU noise and enhance reconstruction performance. We propose an online sequence-level self-consistency strategy (SCS) to improve the reconstruction stability using hierarchical consistency among the generated sub-sequences and the original sequence. The experimental results on the arm and carotid datasets show that our OSCNet achieves state-of-the-art reconstruction performance. Future research will focus on exploring more general reconstruction methods.

\subsubsection*{Acknowledgements}
This work was supported by the grant from National Natural Science Foundation of China (Nos. 62171290, 62101343), Shenzhen-Hong Kong Joint Research Program (No. SGDX20201103095613036), and Shenzhen Science and Technology Innovations Committee (No. 20200812143441001).


%
%

\bibliographystyle{splncs04}
\bibliography{paper1835}

\end{document}